\documentclass[10pt,twocolumn,letterpaper]{article}

\usepackage{cvpr}
\usepackage{times}
\usepackage{epsfig}
\usepackage{graphicx}
\usepackage{amsmath}
\usepackage{amssymb}

\usepackage{subcaption}
\usepackage{booktabs}	
\usepackage{siunitx}
\usepackage[nolist]{acronym}
\begin{acronym}
    \acro{CNN} {Convolutional Neural Network}
    \acro{HOG} {Histogram of Oriented Gradients}
    \acro{ReLU} {Rectified Linear Unit}
    \acro{LCN} {Local Contrast Normalization}
    \acro{NMS} {Non Maxima Suppression}
    \acro{SGD} {Stochastic Gradient Descent}
    \acro{FPPI} {False Positives Per Image}
    \acro{DPM} {Deformable Parts Model}
    \acro{LFPW} {Labeled Face Parts in the Wild}
    \acro{AFLW} {Annotated Facial Landmarks in the Wild}
    \acro{AFW} {Annotated Faces in the Wild}
    \acro{FDDB} {Face Detection Data Set and Benchmark}
    \acro{SSE} {Sum of Squares Error}
    \acro{ACF} {Aggregate Channel Features}
    \acro{AP} {Average Precision}
    \acro{COFW} {Caltech Occluded Faces in the Wild}
    \acro{CCA} {Canonical Correlation Analysis}
    \acro{AUC} {area under the curve}
    \acro{ToF} {Time-of-Flight}
\end{acronym}

\usepackage{color}
\definecolor{green_ours}{rgb}{0.4660, 0.6740, 0.1880}
\definecolor{blue_ours}{rgb}{0, 0.4470, 0.7410}

\usepackage[all]{nowidow}

\makeatletter 
\renewcommand{\paragraph}{%
  \@startsection{paragraph}{4}%
  {\z@}{2ex \@plus 1ex \@minus .2ex}{-1em}%
  {\normalfont\normalsize\bfseries}%
} 
\makeatother

\newcommand\blfootnote[1]{%
  \begingroup
  \renewcommand\thefootnote{}\footnote{#1}%
  \addtocounter{footnote}{-1}%
  \endgroup
}
\usepackage{textcomp}	

\usepackage[hang,flushmargin]{footmisc}	

\usepackage[pagebackref=true,breaklinks=true,letterpaper=true,colorlinks,bookmarks=false]{hyperref}

\cvprfinalcopy 

\def\cvprPaperID{0000} 

\begin{document}

\title{Learning Pose Specific Representations by Predicting Different Views}

\author{Georg Poier \hspace{3em} David Schinagl \hspace{3em} Horst Bischof\\
Institute for Computer Graphics and Vision\\
Graz University of Technology\\
Austria
}

\maketitle

\blfootnote{\textcopyright \,2018 IEEE}	
\blfootnote{Project webpage with code, data and additional material can be found 
at \url{https://poier.github.io/PreView}}

\begin{abstract}
The labeled data required to learn pose estimation for articulated objects 
is difficult to provide in the desired quantity, 
realism, density, and accuracy.
To address this issue, we develop a method to learn representations, 
which are very specific for articulated poses, without the need for labeled training data.
We exploit the observation that the object pose of a known object
is predictive for the appearance in any known view.
That is, given only the pose and shape parameters of a hand, 
the hand's appearance from any viewpoint can be approximated.
To exploit this observation, we train a model that -- given input from one view -- 
estimates a latent representation, which is trained to be predictive
for the appearance of the object when captured from another viewpoint.
Thus, the only necessary supervision is the second view.
The training process of this model reveals an implicit pose representation in the 
latent space.
Importantly, at test time the pose representation can be inferred using only a single view.
In qualitative and quantitative experiments we show that the learned representations
capture detailed pose information.
Moreover, when training the proposed method jointly with labeled and unlabeled data, 
it consistently surpasses the performance of its fully supervised counterpart,
while reducing the amount of needed labeled samples by at least one order of magnitude.

\end{abstract}

\section{Introduction}
In this work we aim to estimate the pose of the hand given a single depth image.
For this task, the best performing methods have recently relied heavily on models learned 
from data~\cite{Guo2017arxiv_goodpractices,Oberweger2015iccv_feedbackloop,
Sun2015cvpr_cascadedhaperegression,Supancic2015iccv_eval}.
Even methods which employ a manually created hand model 
to search for a good fit with the observation,
often employ such a data-driven part as initialization or 
for error correction~\cite{Krejov17cviu,Tang2015iccv_hierarchicalsampling,
Taylor2016siggraph,Ye2016eccv_attentionhybridhape}.
Unfortunately, data-driven models require a large amount of labeled data, 
covering a sufficient part of the pose space, to work well.

However, for the task of estimating the pose of articulated objects, like the human hand, 
it is especially expensive 
to provide accurate annotations for a sufficient amount of real world data.
The articulated structure and specific natural movements of the hand 
frequently cause strong self-occlusions. 
Together with the many 3D points to be annotated, 
this makes the annotation procedure a huge effort for human annotators.

\begin{figure}[t]
  \centering
  \includegraphics[width=\linewidth]{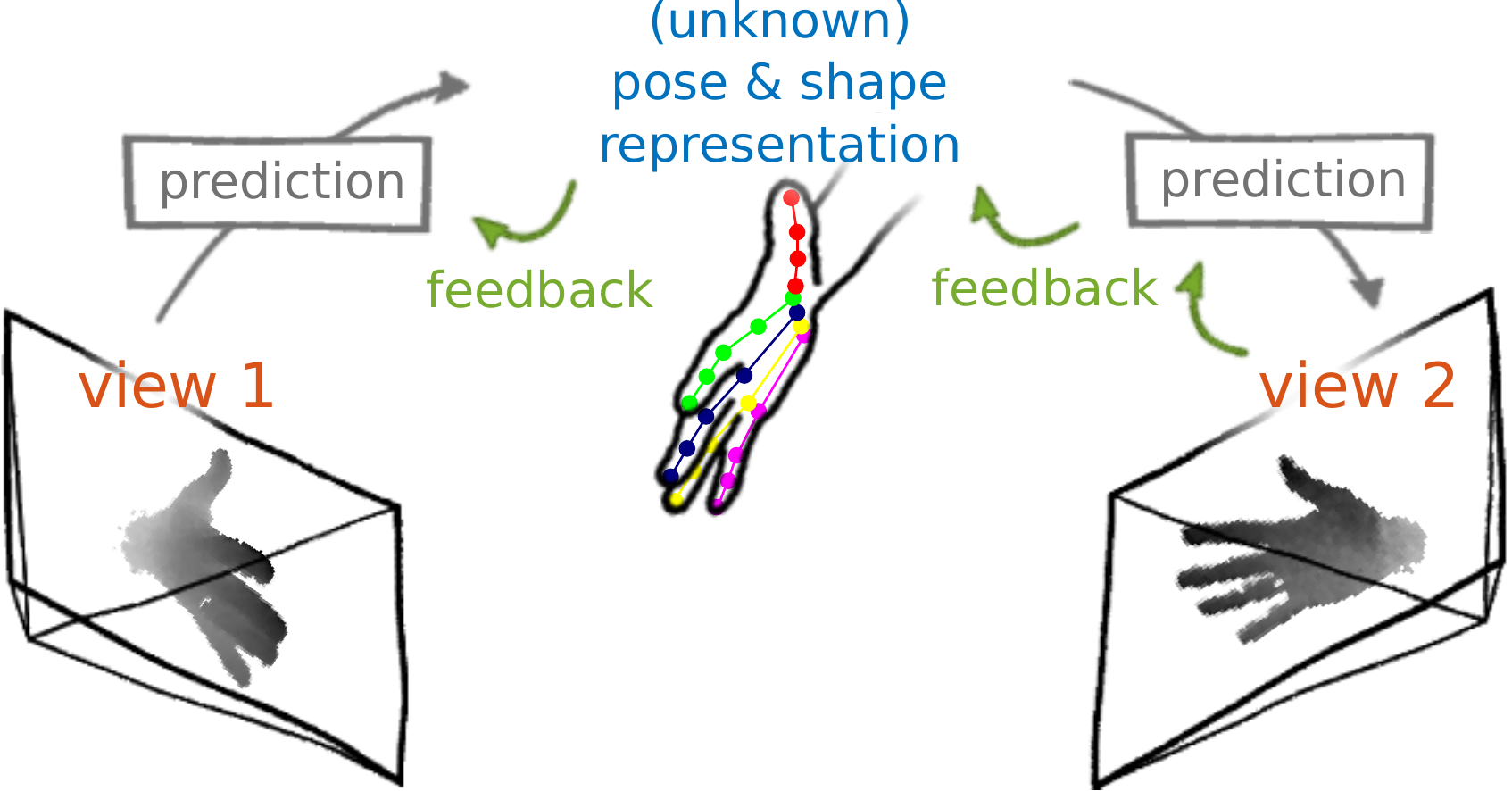}
  \caption[]{\textbf{Sketch for learning a pose specific representation from unlabeled data.}
  We learn to predict a low-dimensional latent representation and, subsequently,
  a different view of the input, 
  \emph{solely} from the \textcolor{blue_ours}{latent representation}.
  The error of the view prediction is used as \textcolor{green_ours}{feedback},
  enforcing the latent representation to capture pose specific information
  without requiring labeled data.
  }
  \label{fig:idea_sketch}
\end{figure}

A largely unexplored direction to cope with this challenge is 
to exploit unlabeled data, which is easy to obtain in large quantities.
We make a step towards closing this gap and 
propose a method that can exploit unlabeled data
by making use of a specific property of the pose estimation task.
We rely on the observation that pose parameters\footnote{For the sake of clarity, here,
\emph{pose parameters} denote the parameters defining the skeleton, 
including its size, as well as a rough shape} 
are predictive for the object appearance of a known object from any viewpoint.
That is, given the pose parameters of a hand, the hand's appearance from any viewpoint
can be estimated.
The observation might not seem helpful upfront,
since it assumes the pose -- which we want to estimate -- to be given.
However, the observation becomes helpful if we capture the scene 
simultaneously from different viewpoints.

By employing a different camera view, we can guide the training of the pose estimation model
(see Fig.~\ref{fig:idea_sketch}).
The guidance relies on the fact that from any set of pose parameters,
which accurately specify the pose and rough shape of the hand,
we necessarily need to be able to predict the hand's appearance in any other view.
Hence, by capturing another view, this additional view can be used as a target
for training a model, which itself guides the training of the underlying pose representation.

More specifically, the idea is to 
train a model which -- given the first camera view --
estimates a small number of latent parameters,
and subsequently predicts a different view solely from these 
few parameters.
The intuition is that the small number of parameters 
resemble a parameterization of the pose. 
By learning to predict a different view from the latent parameters,
the latent parameters are enforced to capture pose specific information.
Framing the problem in this way, a pose representation can be learned just 
by capturing the hand simultaneously from different viewpoints and 
learning to predict one view given the other.

Given the learned low-dimensional pose representation, 
a rather simple mapping to a specific target (\eg, joint positions)
can be learned from a much smaller number of training samples
than required to learn the full mapping from input to target.
Moreover, when training jointly with labeled and unlabeled data,
the whole process can be learned end-to-end in a semi-supervised fashion,
achieving similar performance with one order of magnitude less labeled samples.
Thereby, the joint training regularizes the model to ensure that 
the learned pose representation can be mapped
to the target pose space using the specified mapping.

We show the specificity of the learned representation
and its predictiveness for the pose
in qualitative and quantitative experiments.
Trained in a semi-supervised manner, 
the proposed method consistently outperforms its fully supervised counterpart,
as well as the state-of-the-art in hand pose estimation --
even if all available samples are labeled.
For the more practical case, where the number of unlabeled samples 
is larger than the number of labeled samples,
we find that the proposed method performs on par with the baseline, even with 
one order of magnitude less labeled samples.

\section{Related work}
Traditionally, works on hand pose estimation have been divided into
model-based and data-driven approaches.
Model-based approaches~\cite{Gorce2011pami_modelbasedhape,Melax2013gi_dynamicshape,
Oikonomidis2011bmvc,Roditakis2017bmvc_constrainedsampling,Wu2001iccv_handarticulations} 
search to parameterize a manually created hand model in each frame 
such that it best fits the observation.
These approaches usually need to rely on an initialization, \eg, from previous frames,
and thus, have problems to recover if pose estimation fails once.
Data-driven approaches~\cite{Guo2017arxiv_goodpractices,Keskin2012eccv_multilayeredrf,
Oberweger2017iccvw_deeppriorpp,Tang2014cvpr_lrf}, 
on the other hand, learn a mapping from the input frame to a target pose from a
usually large number of annotated training samples.
These approaches assume that the poses seen at test time 
are at least roughly covered by the training set and will otherwise
fail to deliver a good estimate.
With the desire to combine the merits of both strands, 
researchers have developed hybrid 
approaches~\cite{Mueller2017iccv_realtimeeg,Poier2015bmvc_hybriduncertainties,
Taylor2016siggraph,Ye2016eccv_attentionhybridhape,Zhou16ijcai_modelbaseddeephape}. 
But again, the effectiveness of hybrid approaches is crucially affected by the density 
of the annotations available for training the data-driven part.

\paragraph{Data annotation}
To provide a large number of labeled samples,
(semi-)automatic methods were employed 
to construct the relevant publicly available training sets.
Most often model-based approaches with the above mentioned issues
were used to provide (initial) annotations, which were manually 
corrected~\cite{Sun2015cvpr_cascadedhaperegression,Tang2014cvpr_lrf,Tompson2014tog}.
Other efforts include the development of an annotation 
procedure~\cite{Oberweger2016cvpr_anno} to propagate annotations to similar frames, or 
attaching 6D magnetic sensors to the hand~\cite{Wetzler2015bmvc_fingertip,Yuan2017cvpr_bighand},
which resulted in the largest dataset to date~\cite{Yuan2017cvpr_bighand}.
These efforts underline the difficulties to provide sufficient labeled data, 
hampering novel applications, which might rely on different viewpoints or sensors.

\paragraph{Learning from unlabeled data}
At the same time, capturing unlabeled data is easy, and 
considering the way how we make use of such unlabeled data,
several strands of prior work are related to our method.
The scheme of predicting another view from the learned latent representation 
is, \eg, akin to the concept of autoencoders,
where the input is reconstructed from the latent 
representation~\cite{Hinton2006science_autoenc,Vincent08icml}.
Instead of reconstructing the input, we learn to predict a different view.
This enables the model to capture pose specific representations 
as the results in \S\ref{sec:exp:unsupervised} clearly point out.

Similarly, our work is also related to a strand of works on 
representation learning from unlabeled data
which split the input data into parts and 
have the model learn relations between the 
parts~\cite{Doersch2015iccv_contextprediction,Owens2016eccv_audioselfsupervision,
Pathak2016cvpr_contextencoders,
Zamir2016eccv_3drepresentation,
Zhang2017cvpr_splitbrainautoenc}.
For instance, Doersch~\etal~\cite{Doersch2015iccv_contextprediction} 
learn to predict the relative position of patches sampled from an image,
which should be possible if a model has learned to extract semantics.
Similarly, this has been targeted
by, \eg, relating tracked patches~\cite{Wang15iccv}, 
solving jigsaw puzzles~\cite{Noroozi2016eccv_jigsawpuzzle}
or colorizing images~\cite{Larsson2016eccv}.
While our work can be considered similar in spirit,
our main objective is to learn a pose specific representation in the latent space,
for which a crucial enabler is to employ multiple viewpoints.

\paragraph{Learning from multiple views}
An early example for representation learning from multiple views is 
\ac{CCA}~\cite{Hotelling1936biometrica_cca}
of which various multi-layered, non-linear variants have been 
proposed~\cite{Andrew2013icml_deepcca,Becker1992nature,
Li2016arxiv_multiview,Wang2015icml_deepmultiview_dccae}.
The goal of \ac{CCA} is to relate variables among different views
by learning projections which maximize the correlation between different views.

Researchers have also started to employ multiple camera views 
to learn depth prediction or 3D object reconstruction from unlabeled 
data~\cite{Garg2016eccv_unsuperviseddepth,Jayaraman2017arxiv_shapereconstruction,
Xie2016eccv_deep3d}.
Garg~\etal~\cite{Garg2016eccv_unsuperviseddepth} propose an approach 
to monocular depth estimation, for which the loss is based on the photo consistency 
of the projected pixels in the second view of a stereo image pair.
Similarly, Xie~\etal~\cite{Xie2016eccv_deep3d} target generating a stereo pair 
from a single view. 
Several works add upon this line of research, \eg, 
by incorporating sparse and noisy depth labels~\cite{Kuznietsov2017cvpr_semisuperviseddepth}, 
adding a left-right consistency 
constraint~\cite{Godard2017cvpr_unsuperviseddepthwithlrconsistency},
jointly estimating camera pose and depth~\cite{Zhou2017cvpr_unsupervised},
or learning to reconstruct full 
3D~\cite{Tatarchenko2016eccv_multiviewfromsingleimage,Tulsiani2017cvpr_monocularreconstruction,
Yang2015nips_recurrenttransformsfor3dviewsynth}.

In these works the desired target (\eg, depth or disparity) 
can directly be linked to the training loss via geometric relations
and, therefore, only the intermediate latent representations 
have to encode some kind of semantics of the scene and objects therein.
In our case, the target itself is more explicit semantic (\eg, joint positions or labels, resp.)
and we show how to formulate the task such that our learned latent representation 
closely resembles what we are targeting, namely the pose.
The formulation also clearly differentiates our method from \ac{CCA} and its variants.

\paragraph{Semi-supervised learning for hand pose estimation}
Little work has exploited unlabeled samples for hand pose estimation.
To the best of our knowledge, there are only some notable 
exceptions~\cite{Neverova2015arxiv,Tang2013iccv_semisupervisedhape,
Wan2017cvpr_crossingnets}:
Tang~\etal~\cite{Tang2013iccv_semisupervisedhape} built a discriminative approach
which relies on a large synthetic training set and 
correspondences between synthetic and real samples. 
Similarly, Neverova~\etal~\cite{Neverova2015arxiv} establish correspondences
via an intermediate representation of part segmentations.
For their approach, they do not need pixelwise labels for real samples,
but still require joint annotations.
On the contrary, Wan~\etal~\cite{Wan2017cvpr_crossingnets} incorporate
entirely unlabeled data by drawing from advances 
in generative modeling within a semi-supervised approach.
While elegant and well set up,
neither of these approaches exploit the observation that the pose is 
predictive for the appearance from any known view.

\paragraph{View synthesis for hand pose estimation}
Another notable work on hand pose estimation, we draw inspiration from, 
is the work of Oberweger~\etal~\cite{Oberweger2015iccv_feedbackloop}.
They aim to reconstruct the input view of the hand from 
previously estimated joint positions, and subsequently learn to
generate an update for the pose estimate
based on the discrepancy between the input and the reconstruction
(akin to supervised descent 
methods~\cite{Sheerman2013fg_nonlinearpredictors,Xiong2013cvpr_superviseddescent}).

In contrast to our work, however, they aim to reconstruct the same view 
directly from previous estimates of the joint positions (without capturing shape information).
Consequently, their approach is fully supervised, 
\ie, it requires joint annotations for each sample.
In our work, we do not require pose annotations, 
but exploit the information we get from an additional view point,
which is crucial for the training process, as we will show in our experiments.
Nevertheless, inference is straight forward with our method, \ie, 
we neither require an iterative procedure and generate images  
as in~\cite{Oberweger2015iccv_feedbackloop}, 
nor need a second view at test time.

\section{Learning pose specific representations}
Our work is based on the observation that 
a hand pose representation, $\theta$,
which includes parameters for the hand's size and shape,
is predictive for the hand's appearance, 
$\mathbf{x}^{(i)}$,
from any known view $i$. 
Let $\mathcal{T} \subset \mathbb{R}^{d_{\mathcal{T}}}$ denote the set of possible 
poses or pose representations of dimensionality $d_{\mathcal{T}}$,
\ie, $\theta \in \mathcal{T}$, and similarly,
$\mathcal{X} \subset \mathbb{R}^{d_{\mathcal{X}}}$ be the set of possible input images of
dimensionality $d_{\mathcal{X}}$, \ie, $\mathbf{x}^{(i)} \in \mathcal{X}$.
Then -- based on our observation -- we assume that there exists a view specific mapping, 
$g^{\ast}_{i} \colon \mathbb{R}^{d_{\mathcal{T}}} \rightarrow \mathbb{R}^{d_{\mathcal{X}}}$,
such that
\begin{equation}
 \mathbf{x}^{(i)} = g^{\ast}_{i} ( \theta ), \qquad \forall \theta \in \mathcal{T}.
 \label{equ:posetoappearance}
\end{equation}

Nevertheless, for our task we do not know the pose. 
The pose is what we are searching for.
Given an image of a hand $\mathbf{x}^{(i)}$ we want to find 
the pose of the hand.
That is, we search for a mapping 
$f^{\ast}_{i} \colon \mathbb{R}^{d_{\mathcal{X}}} \rightarrow \mathbb{R}^{d_{\mathcal{T}}}$ 
from the input image to the pose\footnote{To avoid cluttering the notation, we ignore 
that such a mapping is not always unique, given only a single view. 
In theory, we could formulate $\theta$ as a random variable, 
describing a distribution, we could sample from.}:
\begin{equation}
 \theta = f^{\ast}_{i} ( \mathbf{x}^{(i)} ), \qquad \forall \mathbf{x} \in \mathcal{X}.
 \label{equ:appearancetopose}
\end{equation}

Given these two mappings, $f^{\ast}$ and $g^{\ast}$, 
we can see that by subsequently applying them, 
we can directly map from one view to the other.
That is, given an input image of the hand, $\mathbf{x}^{(i)}$,
from view $i$, we can use the mappings to compute 
the hand's appearance $\mathbf{x}^{(j)}$, from any known view $j$:
\begin{equation}
 \mathbf{x}^{(j)} = g^{\ast}_{j}\big( f^{\ast}_{i} ( \mathbf{x}^{(i)} ) \big).
 \label{equ:preview}
\end{equation}

In our case, the mappings $f^{\ast}$ and $g^{\ast}$ are unknown.
We can, however, capture the scene simultaneously from two different views 
$i$ and $j$.
Given the data from two views, $\mathbf{x}^{(i)}$ and $\mathbf{x}^{(j)}$, 
we can formulate our problem as finding a mapping 
from one view to another.
Hence, we use the task of learning a mapping from one view to the other as a ``proxy task''
for finding a latent representation, which resembles the pose.

Note, for $i = j$, Eqn.~\eqref{equ:preview} 
essentially specifies an autoencoder.
In this case it is difficult to ensure that the model 
learns a latent representation, which resembles the pose.
Hence, the crucial case, which we are investigating in this work, 
is the case $i \neq j$.
From our empirical investigation (see~\S\ref{sec:exp:unsupervised}) we find that 
for $i \neq j$ and a sufficient amount of (unlabeled) data
it is easy to constrain the model such that the latent representation
captures pose information.

\subsection{Implementation of the observations}
\label{sec:met:unlabeled}

To implement our observations we want to learn the two mappings, 
$f^{\ast}$ and $g^{\ast}$, from data.
We do so by employing a \ac{CNN} with an encoder-decoder architecture.
To formalize our method, 
we denote the learned estimates of the ``true'' mappings $f^{\ast}$ and $g^{\ast}$, 
$f$ and $g$, respectively.
The encoder $f_{i}$ receives input $\mathbf{x}^{(i)}$ from view $i$ 
and its output represents the desired latent representation $\theta$.
The latent representation is at the same time the input for the decoder $g_{j}$, 
which produces the view $\mathbf{x}^{(j)}$ given $\theta$.
Without loss of generality we assume the captured images, $\mathbf{x}$, to be depth images.
Note that, while, for color-only input the appearance is affected by
additional factors like skin color or illumination, 
the basic observations still hold.

In the basic model, we train our system to predict a 
different view $\mathbf{x}^{(j)}$, which we capture for training.
The training loss, $\ell_{\mathrm{u}}$, for this model can thus be formulated as 
a reconstruction loss
\begin{equation}
 \ell_{\mathrm{u}} = \ell_{\mathrm{recon}} \big( \mathbf{\hat{y}}^{(j)}, \mathbf{x}^{(j)} \big),
\end{equation}
where $\mathbf{\hat{y}}^{(j)}$ is the model's prediction for view $j$, given input 
$\mathbf{x}^{(i)}$ from 
view $i$, \ie,
\begin{equation}
 \mathbf{\hat{y}}^{(j)} = g_{j}\big( f_{i} ( \mathbf{x}^{(i)} ) \big).
\end{equation}
For the reconstruction loss $\ell_{\mathrm{recon}}$ the $L^{1}$ norm 
yielded the best results in our experiments.

Ideally, we want the latent representation, $\theta = f_{i} ( \mathbf{x}^{(i)} )$,
to be very specific for the pose, 
not capturing any unnecessary information.
The loss itself does not constrain the latent representation
to fulfill such a requirement.
We can, however, constrain the latent representation 
in a very simple -- though effective -- way:
We assume that the smallest possible representation which is 
predictive for the appearance of any known view, other than the input,
will, crucially, contain a representation resembling the pose.

A low-dimensional representation of the pose is often given by the joint positions.
However, since there are many dependencies between the joints, 
the pose can even be represented by a lower-dimensional subspace.
While works on hand modeling~\cite{Albrecht03handmodels,Lin2000_handconstraints}
give an indication for the size of such a low-dimensional subspace,
we investigate the size best matching our requirements 
in the experimental section (\S\ref{sec:exp:unsupervised}).
The representation should contain only little additional information 
which could obfuscate the pose representation and, 
thus, hamper learning a mapping to any target pose representation 
as discussed in the next section.

\subsection{Learning from labeled and unlabeled data}
\label{sec:met:semisupervised}
To map from the latent representation space to the desired target space (\eg, joint positions) 
we add a single linear layer to our encoder-decoder architecture.
We enforce the latent representation to suffice this linear map
by training the encoder, which maps from input to the latent representation, 
jointly with labeled and unlabeled data in a semi-supervised manner.
That is, labeled samples guide the training of the latent representation 
such that it suffices the linear mapping.

\begin{figure}[t]
  \centering
  \includegraphics[width=0.99\linewidth]{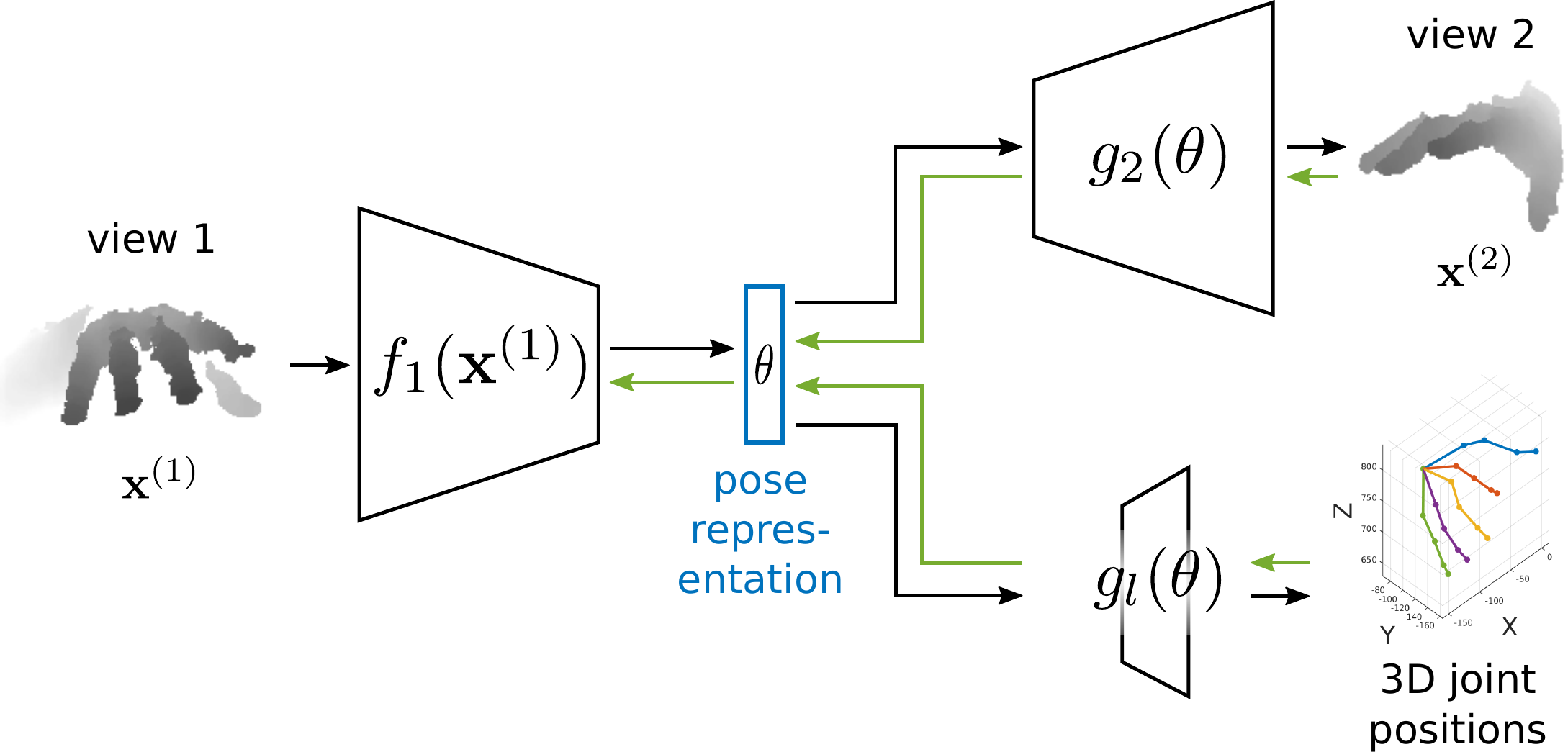}
  \caption[]{\textbf{Architecture sketch for semi-supervised learning.}
  The input view, $\mathbf{x}^{(1)}$, is mapped to the 
  \textcolor{blue_ours}{latent representation}, $\theta$, 
  by the encoder $f_{1}$.
  Based on $\theta$, 
  the decoder $g_{2}$ is required to generate a different view $2$ of the input.
  At the same time the latent representation is ensured to suffice a 
  linear mapping, $g_{l}$, to the 3D joint positions by employing labeled samples.
  This is illustrated by the green paths depicting the \textcolor{green_ours}{gradient flow} 
  to the latent representation and, consequently, to the encoder.}
  \label{fig:semisupervisedarchitecture}
\end{figure}

The architecture for semi-supervised training is depicted in 
Fig.~\ref{fig:semisupervisedarchitecture}.
The parameters of the linear layer from the latent pose representation 
to the joint positions are only trained using labeled samples. 
All other parameters are trained using both labeled and unlabeled samples.

The semi-supervised loss function, $\ell_{\mathrm{semi}}$, 
is a combination of the loss from unlabeled and labeled data:
\begin{equation}
  \ell_{\mathrm{semi}} = \ell_{\mathrm{u}} + \lambda_{\mathrm{l}} \, \ell_{\mathrm{l}},
  \label{equ:loss_semi}
\end{equation}
where $\lambda_{\textrm{l}}$ is a weighting factor, which is set to zero for unlabeled samples.
For robustness, we employ the sum of the Huber loss~\cite{Huber1964annals_loss}
for individual joint errors, which is different from the standard use of the Huber loss.
That is, 
\begin{equation}
  \ell_{\mathrm{l}} = \sum_{k} \ell_{\mathrm{Huber}}
  \left( \left\| y^{\mathcal{J}}_{k} - \hat{y}^{\mathcal{J}}_{k} \right\|_{2} \right),
\end{equation}
where $\left\| . \right\|_{2}$ denotes the $L^{2}$-norm, 
$\hat{y}^{\mathcal{J}}_{k}$ the estimated position of the $k$-th joint, 
$y^{\mathcal{J}}_{k}$ the corresponding ground truth position and
\begin{equation}
 \ell_{\mathrm{Huber}} (d) = 
 \begin{cases}
  0.5 \, d^{2} 		& \textrm{if} \; d < \epsilon \\
  \epsilon \left( d - 0.5 \, \epsilon \right) 	& \textrm{otherwise}.
 \end{cases}
\end{equation}
Note that $d$, the input to the Huber loss, is always positive in our case.

Additionally, an adversarial loss can be added 
to the training objective~\cite{Goodfellow2014nips_gan,Mao2017iccv_lsgan,Mirza2014arxiv_condgan}.
We describe this approach and experiments in the appendix~(\S\ref{sup:sec:adversarial}), 
but omit it from the main paper since 
results are only slightly improved in some cases,
whereas the adversarial loss imposes a computational overhead and requires
a very sensible adjustment of meta-parameters for training.

\subsection{Implementation details}
Similar to other works~\cite{Krejov17cviu,Tang2014cvpr_lrf} 
we assume the hand to be the closest object to the camera, and 
compute its center of mass (CoM),
which is also provided as additional input to the decoder, $g$.
We then crop a region with equal side length in each direction around the CoM, 
resize it to $64 \times 64$ pixels and normalize the depth values within a fixed range
to be between $-1$ and $1$.
These crops form the input to our method.

Our method does not rely on a specific choice of the network architecture.
For our experiments, we implemented our encoder and decoder 
networks based on the architecture developed for DCGAN~\cite{Radford2016iclr_dcgan},
since it is a well developed architecture, which is comparably ``lightweight'' and 
designed for image synthesis.
We base our encoder $f$ on the discriminator and our decoder $g$ on the generator of 
the original publicly available 
implementation.
We only interchange the positions of the ReLUs~\cite{Fukushima1980biolcybernetics_neocognitron} 
and leakyReLUs~\cite{Maas13icml_rectifiernonlinearities} since we want 
to ease gradient flow through the decoder, put a hyperbolic tangent (Tanh) activation 
function at the end of the decoder to ensure that the output can range between $-1$ and $1$,
and adapt the input and output dimensions accordingly.

We train our model with \emph{Adam}~\cite{Kingma2015iclr_adam} for 100 epochs
using a batch size of 128 and a learning rate of $10^{-4}$.
For semi-supervised learning we obtained the best results with $\lambda_{\mathrm{l}}~=~10$.
Our \emph{PyTorch} implementation is publicly 
available\footnote{\url{https://poier.github.io/PreView}}
\addtocounter{footnote}{-1}\addtocounter{Hfootnote}{-1}.

\section{Experiments}
To prove the applicability of the proposed method
we perform qualitative and quantitative experiments on different datasets.
We investigate the representations learned from unlabeled data (\cf~\S\ref{sec:met:unlabeled})
in \S\ref{sec:exp:unsupervised}. 
Subsequently, we present the results for 
semi-supervised learning (\cf~\S\ref{sec:met:semisupervised}),
compare to the state-of-the-art in hand pose estimation, and
provide evidence for the effectiveness of our training procedure in an ablation study 
(\S\ref{sec:exp:semisupervised}).

\subsection{Datasets}
\label{sec:exp:datasets}

We evaluate on two different datasets.
Firstly, we test on the NYU hand pose dataset~\cite{Tompson2014tog},
which, to the best of our knowledge, is the only public dataset 
providing multiple views for the training and test set.
For a broader empirical analysis of our approach we additionally 
provide a novel multi-view dataset\footnotemark.

\paragraph{NYU hand pose dataset}
The NYU dataset provides a training set with 72,757 frames from a single actor and 
a test set with 8,252 frames from two actors.
It was captured with structured light based RGBD cameras.
The additional cameras captured the scene from side views.
Originally, the additional cameras were employed to mitigate issues with 
self-occlusions during annotation; for our work the additional camera views
enable us to compare our approach on a standard dataset.
Unfortunately, the side view camera locations were changed several times 
during training set acquisition and no camera pose information is provided.
Therefore, we searched for a part of the training set with approximately 
similar camera setup and found 43,641 frames ($\sim$60\% of the original training set),
which we used as a training set for our experiments.
For validation and testing, we use the full sets from the original dataset.
We denote the reduced training set with consistent setup by \emph{NYU-CS}.

\paragraph{Multi-view hand pose dataset}
We captured the dataset for typical user interaction scenarios in front of 
a large screen with a \acl{ToF} camera mounted at each of the two top corners of the screen.
We captured the two cameras synchronously and 
captured poses needed for typical gestures like swiping, pointing or waving.
While the set of poses is restricted, we aimed to capture each pose in all possible 
hand orientations and ended up with 63,701 frames from 14 different actors.
Since the goal of our novel dataset is to investigate semi-supervised learning 
where only a small fraction of the available samples is labeled,
we only labeled a representative subset from a few actors. 
To this we employed the method in~\cite{Oberweger2016cvpr_anno},
which tries to find a subset of frames covering the pose space well.
Overall 526 frames from 7 out of the 14 actors were manually annotated.
We split the labeled data in 289 frames for training and validation (189/100) and 
237 for testing.
We denote the resulting multi-view hand pose dataset \emph{MV-hands}.

\subsection{Metrics}
\label{sec:exp:metrics}
For the evaluation, we employ three commonly used metrics:
the mean joint error (\emph{ME}) as well as the joint- and 
frame-based success rate (\emph{JS}/\emph{FS}).
The ME denotes the average distance between the estimated and 
ground truth joint positions in millimeter (\si{\milli\meter}).
The JS is the fraction of joints which were 
estimated within a certain distance to the ground truth joint 
position.
The FS is stricter and gives the fraction of frames 
for which all joints have been estimated within a certain distance to the ground truth
position~\cite{Taylor2012cvpr_vitruvian}.

For hand pose estimation researchers often employ
curves of the success rates over different distance thresholds.
To express these curves with a single number, we compute the \ac{AUC} 
up to a specified threshold.
We denote the \ac{AUC} of the JS and FS 
up to a distance threshold of \SI{80}{\milli\meter} 
by \emph{JS80}, and \emph{FS80}, respectively.

\subsection{Pre-training from unlabeled data}
\label{sec:exp:unsupervised}

In the following, we perform several experiments 
to investigate the effectiveness of representations learned from unlabeled data. 

\begin{table}[t]
\begin{center}
\begin{tabular}{l c c c}
\toprule
$n$ & Autoencoder & \textbf{PreView (Ours)} \\ 
\midrule
100 & $48.0 \pm 0.76$ & $\mathbf{33.4} \pm 1.18$ & $-30.4\%$ \\ 
1,000 & $47.2 \pm 0.29$ & $\mathbf{29.6} \pm 0.32$ & $-37.3\%$ \\ 
10,000 & $47.3 \pm 0.08$ & $\mathbf{29.0} \pm 0.14$ & $-38.7\%$ \\ 
43,640 & $47.1 \pm 0.08$ & $\mathbf{29.0} \pm 0.09$ & $-38.4\%$ \\ 
\bottomrule
\end{tabular}
\end{center}
\caption{
  \textbf{Pre-training from unlabeled data.}
  Mean joint error and standard deviation on the NYU-CS dataset 
  for different pre-training methods and numbers of labeled samples, $n$.
  }
\label{tab:unsupervised}
\end{table}

\paragraph{Linear mapping to joint positions}
\label{sec:exp:pretrain}
To quantitatively analyze the predictability of the pose 
given the learned latent representations,
we follow the standard procedure for testing representations learned in an un-/self-supervised 
manner~\cite{Coates2011aistats_singlelayerunsupervised,Dosovitskiy14nips,
Noroozi2017iccv_learntocount,Zhang2016eccv_colorful}:
We train the network using the respective pre-training method, 
\ie, without pose annotations, 
freeze all layers up to the latent pose representation 
and train a linear mapping from the latent representation 
to the target joint positions using annotated samples.

The results on the NYU-CS dataset are shown in Tab.~\ref{tab:unsupervised}.
We compare our method to pre-training using an autoencoder because of its close relation.
In particular, the autoencoder's target is the input view, whereas 
our method aims to predict a different view.
For a fair comparison, we use the same architecture, \ie, 
the same number of parameters and training algorithm
for the autoencoder and the proposed method for predicting different views (\emph{PreView}).

Here, we also investigate how the respective methods behave 
when the number of labeled samples, $n$, is smaller than the number of unlabeled samples,
\ie, only a subset of labeled samples is provided.
In this case, we use a random subset of the data, which is the same for each method.
For the case where the training set is small,
the size of the validation set will -- for a realistic scenario -- be similarly small.
To account for this, we also subsample the validation data.
We fix the size of the validation set, 
$\left| \mathcal{V} \right|$, as a fraction of
the size of the sub-sampled training set, $\left| \mathcal{L} \right|$. 
That is, we sample at most 
$\left| \mathcal{V} \right| \leq 0.3 \left| \mathcal{L} \right|$ samples 
from the original validation set.
We repeat this experiment 10 times with different random samples
to investigate the effect of the sampling and 
report the average and standard deviation of the results in Tab.~\ref{tab:unsupervised}.

The results show that pre-training for view prediction yields a 
latent representation which is significantly more predictive for the pose 
than pre-training using an autoencoder.
The improvement is consistent 
-- independent of the ratio between labeled and unlabeled samples -- 
and ranges between 30 and 40 percent.

On the other hand, qualitative inspection shows that 
the autoencoder yields cleaner reconstructions 
of the inputs, compared to the predictions of the second view of our method.
Obviously, reconstructing the input is an easier task, and can be done 
more accurately, even without knowledge about the pose, as the results  
in Tab.~\ref{tab:unsupervised} suggest.
We show some exemplary view predictions 
and input reconstructions in the appendix~(\S\ref{sup:sec:predictedviews}).

We believe that there are several reasons for this large improvement in pose predictability: 
For example, our model is enforced to not just capture pixel statistics 
as can be sufficient to reconstruct the 
input~\cite{Hotelling1933jedupsy_pca,Kirby1990pami_eigenfaces,Pearson1901philmag_pca}
since the prediction of a different viewpoint requires the model to actually 
reason about the pose. 
More specifically, our model needs to reason about how the appearance affects the pose 
and thus the appearance in the other view.

\begin{figure}[t]
  \centering
  \includegraphics[width=\linewidth]{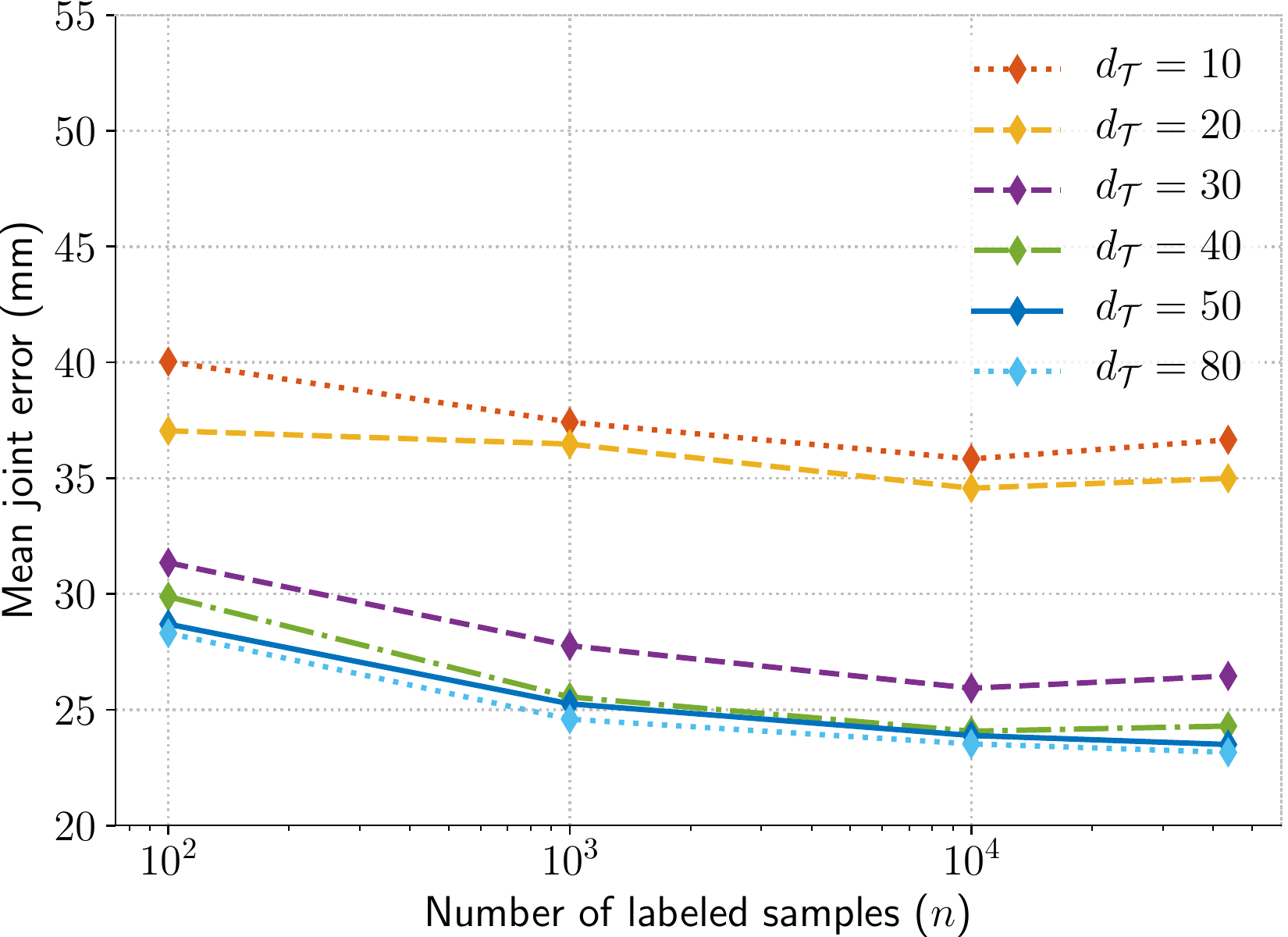}
  \caption[]{\textbf{Pose predictability.} 
  How the size of the latent representation, $d_{\mathcal{T}}$, 
  affects the predictability of the pose (from pre-trained, frozen representations). 
  Results on the NYU validation set.}
  \label{fig:embeddingsize}
\end{figure}

\begin{figure*}[t]
  \centering
  \begin{subfigure}{0.48\linewidth}
    \includegraphics[width=\textwidth]{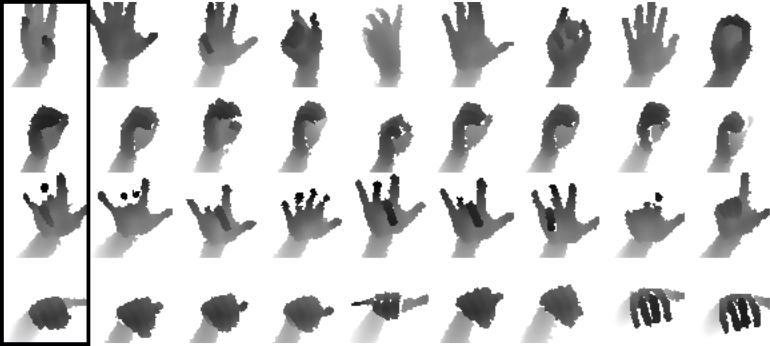}
    \caption{Autoencoder}
    \label{fig:nearestneighbors:autoenc}
  \end{subfigure}
  \quad
  \begin{subfigure}{0.48\linewidth}
    \includegraphics[width=\textwidth]{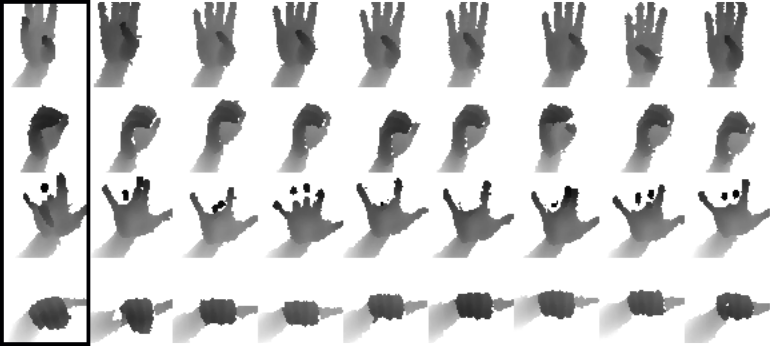}
    \caption{PreView (Ours)}
    \label{fig:nearestneighbors:preview}
  \end{subfigure}
  \caption{\textbf{Nearest neighbors in latent space.}
  Comparison of nearest neighbors in the latent representation space 
  for representations learned using an autoencoder (a) and our method (b).
  Query images (same queries shown for both methods) -- 
  randomly sampled from the validation set -- 
  are shown in the marked, leftmost column of (a) and (b), 
  the remaining columns are the respective nearest neighbors.
  }
  \label{fig:nearestneighbors}
\end{figure*}

\paragraph{Size of the latent representation}
We expect the size of the latent representation to be an important constraint for
the specificity of the learned pose 
representation (\cf~\cite{Oberweger2015cvww_deepprior,Tekin2016bmvc_structuredpose}).
Hence, we investigate how the size of the 
representation affects the results, 
\ie, the predictability of the pose.
For this hyperparameter evaluation we employ the NYU validation set.
We compare the results for representations of size 
$d_{\mathcal{T}}~\in~\{10, 20, 30, 40, 50, 80 \}$ in Fig.~\ref{fig:embeddingsize}.
It shows that the mean joint error is reduced by a large margin when increasing 
$d_{\mathcal{T}}$ from $20$ to $30$,
but the improvement diminishes 
if $d_{\mathcal{T}} \sim 40$.
It seems that, when trained in the proposed way, 
a size of $20$ and below is too small to capture the 
pose and shape parameters reasonably well.
However, if the size of the representation is increased above $50$ 
the predictability of the pose is not improved anymore.
This is interesting, since the size of the parameter space,
which was identified by works on hand 
modeling~\cite{Albrecht03handmodels,Lin2000_handconstraints}
is usually very similar. 
The size identified in these works is indeed slightly smaller when representing the pose alone.
In our case, however, the learned latent representation also 
needs to capture the size and shape of the hand.

\paragraph{Neuron activations}
We also aimed to investigate what each neuron in the latent space has learned.
When we search for the samples from the validation set, 
which activate a single neuron most,
we can observe that many of the neurons are activated most for very specific poses.
That is, the samples, which activate a neuron most, clearly show similar poses.
We include these qualitative samples in the appendix~(\S\ref{sup:sec:neuronactivations}).

\paragraph{Nearest neighbors}
To obtain further insights into the learned representation, 
we visualize nearest neighbors in the latent representation space.
More specifically, given a query image from the validation set, we find the closest samples 
from the training set according to the Euclidean distance in the latent representation space.
Fig.~\ref{fig:nearestneighbors:preview} visualizes some randomly sampled query images 
(\ie, no ``cherry picking``) and their corresponding nearest neighbors.
We see that the nearest neighbors most often exhibit a very similar pose as the 
query image, even if the detection (\ie, hand crop) is not always accurate.
This is in contrast to the nearest neighbors in the latent representation learned 
using autoencoders, which often show a completely different pose 
(see Fig.~\ref{fig:nearestneighbors:autoenc}).

\subsection{Semi-supervised training}
\label{sec:exp:semisupervised}
In a final set of experiments we test the proposed method for jointly leveraging 
labeled and unlabeled data (\cf~\S\ref{sec:met:semisupervised}) 
during end-to-end training.
Similar to the previous setup, 
we consider the case where the number of labeled samples 
is smaller or equal than the number of unlabeled samples, and 
evaluate different ratios.
For a small number of labeled samples 
we obtained the best results by sampling the mini-batches such that 
there is an equal amount of labeled and unlabeled samples in each batch 
(\cf,~\cite{Zhou2017iccv_3dposeinthewild}).

\begin{table*}[t]
\small
\setlength{\tabcolsep}{0.59em}
\begin{center}
\begin{tabular}{l c c c c c c c c c c c c} 
\toprule
$n$ & \multicolumn{3}{c}{100} & \multicolumn{3}{c}{1,000} & \multicolumn{3}{c}{10,000} & \multicolumn{3}{c}{43,640} \\ 
\cmidrule(rl){2-4} \cmidrule(rl){5-7} \cmidrule(rl){8-10} \cmidrule(rl){11-13}
Metric (see~\S\ref{sec:exp:metrics}) & ME & FS80 & JS80 & ME & FS80 & JS80 & ME & FS80 & JS80 & ME & FS80 & JS80 \\ 
\midrule
DeepPrior~\cite{Oberweger2015cvww_deepprior} & $44.99$ & $0.11$ & $0.45$ & $36.99$ & $0.20$ & $0.55$ & $30.31$ & $0.31$ & $0.63$ & $27.97$ & $0.35$ & $0.66$ \\ 
Crossing Nets~\cite{Wan2017cvpr_crossingnets} & $67.65$ & $0.00$ & $0.25$ & $36.35$ & $0.16$ & $0.55$ & $28.97$ & $0.29$ & $0.64$ & $25.57$ & $0.34$ & $0.68$ \\ 
DeepPrior++~\cite{Oberweger2017iccvw_deeppriorpp} & $38.07$ & $0.14$ & $0.53$ & $31.01$ & $0.23$ & $0.61$ & $24.14$ & $0.37$ & $0.69$ & $20.87$ & $0.44$ & $0.73$ \\ 
\midrule
Semi-superv. Autoenc. & $31.58$ & $0.27$ & $0.60$ & $24.05$ & $0.41$ & $0.70$ & $21.32$ & $0.47$ & $0.73$ & $20.74$ & $0.49$ & $0.74$ \\ 
\textbf{Semi-superv. PreView (Ours)} & $\mathbf{29.35}$ & $\mathbf{0.31}$ & $\mathbf{0.63}$ & $\mathbf{22.83}$ & $\mathbf{0.43}$ & $\mathbf{0.71}$ & $\mathbf{19.81}$ & $\mathbf{0.50}$ & $\mathbf{0.75}$ & $\mathbf{19.60}$ & $\mathbf{0.51}$ & $\mathbf{0.75}$ \\ 
\bottomrule
\end{tabular}
\end{center}
\caption{\textbf{Comparison to the state-of-the-art.}
  Results on the NYU-CS dataset for different metrics and 
  different numbers of labeled samples $n$.
  For the mean joint error (ME) smaller values are better, while for the 
  success rates (FS80 and JS80) higher values are better. 
  Best results in boldface.
  }
\label{tab:exp:sota}
\end{table*}

\begin{table}[t]
\setlength{\tabcolsep}{0.57em}
\begin{center}
\begin{tabular}{l c c c}
\toprule 
$n$ & \multicolumn{3}{c}{289} \\ 
\cmidrule(rl){2-4}
Metric (see~\S\ref{sec:exp:metrics}) & ME & FS80 & JS80 \\ 
\midrule
DeepPrior++~\cite{Oberweger2017iccvw_deeppriorpp} & $34.17$ & $0.22$ & $0.57$ \\ 
\midrule
Supervised & $26.35$ & $0.36$ & $0.67$ \\ 
Semi-superv. Autoencoder & $25.20$ & $0.38$ & $0.68$ \\ 
\textbf{Semi-superv. PreView (Ours)} & $\mathbf{24.14}$ & $\mathbf{0.39}$ & $\mathbf{0.69}$ \\
\bottomrule
\end{tabular}
\end{center}
\caption{\textbf{Comparison to the state-of-the-art and ablation experiments.}
  Results for different metrics on the MV-hands dataset.
  }
\label{tab:exp:semisupervised:ablation:icg:289}
\end{table}

\paragraph{Comparison to the state-of-the-art}
To evaluate the competitiveness of the employed architecture, 
we compare against the state-of-the-art in data-driven hand pose estimation.
Since the NYU-CS set contains about 60\% of the original training set,
we need to re-train the state-of-the-art approaches on the same subset for a fair comparison.
We compare to Crossing Nets~\cite{Wan2017cvpr_crossingnets}, 
DeepPrior~\cite{Oberweger2015cvww_deepprior} and 
DeepPrior++~\cite{Oberweger2017iccvw_deeppriorpp}.
We selected DeepPrior, since its 
results are still in the range of the state-of-the-art
for the NYU dataset 
(as shown in a recent independent evaluation~\cite{Yuan2017cvpr_bighand}),
the PCA based ''prior`` makes the approach suffer less 
from a reduced training set, and finally,
it has about the same number of model parameters as our model.
The improved variant DeepPrior++, on the other hand, 
has very recently been shown to be top-performing 
on different datasets~\cite{Oberweger2017iccvw_deeppriorpp}.

To train the state-of-the-art approaches, we use the publicly available 
source code provided by the authors.
Note, Wan~\etal~\cite{Wan2017cvpr_crossingnets} used
different models for the experiments on the NYU dataset 
than the ones used in their publicly available code.
For a fair comparison we use the same (metric) crop size 
when cropping the hand for the entire training and test set, and
fix the training and validation subsets to the same subsets as for the evaluation 
of our method.

The results in Tab.~\ref{tab:exp:sota} and~\ref{tab:exp:semisupervised:ablation:icg:289}
show that -- by leveraging unlabeled data -- 
our method consistently improves the performance, independent of the number of labeled samples,
and improves the state-of-the-art approaches by a large margin 
for a small number of labeled samples.
Note that the NYU dataset does not provide additional unlabeled samples, \ie,
when all labeled samples are used, our method can not draw from any additional 
information.

\paragraph{Ablation experiments}
Finally, we focus the quantitative evaluation on the main contribution of this work.
We exclude disturbing factors like the model architecture 
or the training procedure
by training a baseline for which we keep everything the same 
but do not exploit any unlabeled data.

\begin{figure}[t]
  \centering
  \includegraphics[width=\linewidth]{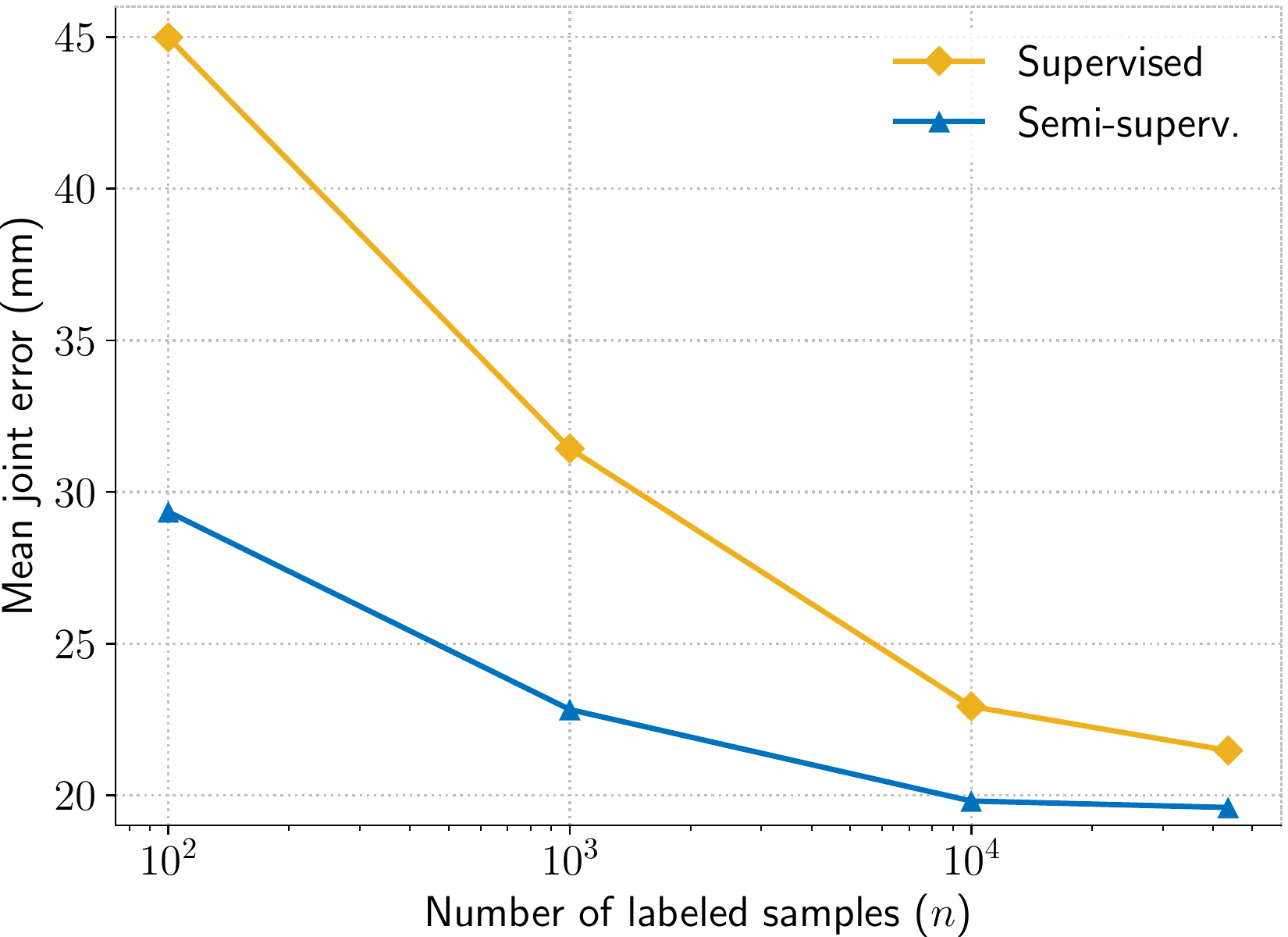}
  \caption[]{\textbf{Ablation experiments.}
  Comparison of purely supervised training (\emph{Supervised}), 
  with the proposed method which can exploit unlabeled samples (\emph{Semi-superv.})
  for different numbers of labeled samples $n$ on the NYU-CS dataset.
  }
  \label{fig:exp:semisupervised:ablation:nyu}
\end{figure}

In Tab.~\ref{tab:exp:semisupervised:ablation:icg:289} we compare the results on the MV-hands dataset.
We see that our semi-supervised training 
improves the results of supervised training for all metrics.
Fig.~\ref{fig:exp:semisupervised:ablation:nyu} compares the results on the NYU-CS dataset,
where our method (\emph{Semi-superv.}) also
improves results for a high number of labeled samples.
For the more realistic case, where only a subset of the data is labeled, 
our method improves the performance of the fully supervised approach by a large margin.
In fact, our method achieves similar or improved results 
even when it is trained with one to two orders of magnitude less labeled samples.

\section{Conclusion}
Learning from unlabeled data has long been recognized as an important direction for 
machine learning and appears especially desirable for tasks with high labeling effort, 
such as estimation of articulated poses.
However, traditionally the representations learned from unlabeled data are most often generic.
While in this way the representations are amenable for transfer learning to novel tasks,
concrete applications benefit from task specific representations. 

In this work, we showed a way how to learn task specific representations
for pose estimation without labels.
Additionally, the proposed method can be trained end-to-end in a semi-supervised manner.
Our method consistently surpasses the performance of standard supervised training, 
even when all available training samples are labeled.
Moreover, the results of supervised training are already improved with 
one order of magnitude less labeled training samples.

{
\paragraph{Acknowledgements}
We thank the anonymous reviewers for their effort and valuable feedback, 
Samuel Schulter, Peter Roth, Michael Opitz and Horst Possegger 
for feedback on an early draft of this manuscript, 
Chengde Wan for his responsiveness regarding their implementation 
of~\cite{Wan2017cvpr_crossingnets},
Martin Oswald, Andreas Bean and colleagues at IGT Unterpremst\"atten 
for the fruitful cooperation and lending their hands for the dataset, and
Anna Micheler-Hofer, Stefan Ainetter and Florian Ziessler 
for manually labeling hand poses.
We also thank the NVIDIA Corporation for awarding a Titan Xp GPU 
without which less experiments would have been possible.
}

{\small
\bibliographystyle{ieee}
\bibliography{../_abbrv_short,../references}
}

\clearpage
\pagenumbering{roman}
\setcounter{page}{1}
\appendix

\section{Appendix}
This appendix provides additional results 
supporting the claims of the main paper.
First, we provide details about the mentioned adversarial training and
compare the respective results quantitatively (\S\ref{sup:sec:adversarial}).
Additionally, we compare to a more recent pre-training 
baseline~(\S\ref{sec:sup:contextencodercomparison}) and 
provide the results when omitting the validation set for training on our novel 
dataset~(\S\ref{sec:sup:novalmvhands}).
Finally, we show more examples from our qualitative investigation.
Specifically, we show examples for predicted views~(\S\ref{sup:sec:predictedviews}),
investigate which inputs activate the neurons in the latent representation 
most~(\S\ref{sup:sec:neuronactivations}), 
and visualize the nearest neighbors in the latent space for our novel 
dataset~(\S\ref{sup:sec:nearestneighbors}).

\subsection{Adversarial training}
\label{sup:sec:adversarial}
In~\S\ref{sec:met:semisupervised} we stated that we also experimented with an 
additional adversarial loss term. 
Here, we give some details about the underlying intentions, 
the implementation and the results of these experiments.

In our work, the objective for the decoder $g$ is based on the reconstruction loss.
In this way, the decoder is penalized for any deviation 
from the second view's exact pixel values.
However, more important for our task is the global structure of the image
as this is affected crucially by the pose.
That is, the decoder spends representational power on estimating exact pixel values,  
which are of little interest to us.

The adversarial training procedure, on the other hand, 
as proposed by Goodfellow~\etal~\cite{Goodfellow2014nips_gan}
corresponds to a minimax two-player game, where each player is implemented by a neural network.
A generator network aims to generate samples from the data distribution and a 
discriminator network aims to discriminate generated samples from real samples.
In this game, the loss for the generator is essentially provided by the discriminator,
thus overcoming the need for explicit supervision, \eg, from corresponding target images.

Using this idea, we can train the decoder $g$ of our method to match the distribution of 
real images, but lessen the focus on raw pixel differences. 
We do so by adding an additional adversarial term 
to the loss in Eq.~\eqref{equ:loss_semi},
\begin{equation}
  \ell_{\mathrm{semi}} = \ell_{\mathrm{u}} + \lambda_{\mathrm{l}} \, \ell_{\mathrm{l}}
			+ \lambda_{\mathrm{a}} \, \ell_{\mathrm{a}},
\end{equation}
where $\lambda_{\mathrm{a}}$ is a weighting factor and 
$\ell_{\mathrm{a}}$ is based on how ``real'' the discriminator network $h$ 
thinks a generated sample $\mathbf{\hat{y}}$ is.
That is, since this yielded the best results,
we define $\ell_{\mathrm{a}}$ inspired by Least Squares GAN~\cite{Mao2017iccv_lsgan} as
\begin{equation}
 \ell_{\mathrm{a}} = \frac{1}{2} \left( h_{j} ( \mathbf{\hat{y}}^{(j)} ) - l_{r} \right)^{2},
\end{equation}
where $l_{r}$ is the label value for real samples.
The objective for the discriminator, on the other hand, is to push 
the real samples towards $l_{r}$ and 
generated samples towards a distinct label value $l_{g}$, \ie,
\begin{equation}
 \ell_{\mathrm{h}} = \frac{1}{2} \left( h_{j} ( \mathbf{x}^{(j)} ) - l_{r} \right)^{2}
		   + \frac{1}{2} \left( h_{j} ( \mathbf{\hat{y}}^{(j)} ) - l_{g} \right)^{2}.
\end{equation}
In our case we set $l_{r} = 1$ and $l_{g} = 0$.
For the adversarial part, we adopted the training procedure and discriminator architecture of 
DCGAN~\cite{Radford2016iclr_dcgan}.

\begin{table*}[t]
\small
\begin{center}
\begin{tabular}{l c c c c c c c c c c c c}
\toprule
$n$ & \multicolumn{3}{c}{100} & \multicolumn{3}{c}{1,000} & \multicolumn{3}{c}{10,000} & \multicolumn{3}{c}{43,640} \\ 
\cmidrule(rl){2-4} \cmidrule(rl){5-7} \cmidrule(rl){8-10} \cmidrule(rl){11-13}
Metric (see~\S\ref{sec:exp:metrics}) & ME & FS80 & JS80 & ME & FS80 & JS80 & ME & FS80 & JS80 & ME & FS80 & JS80 \\ 
\midrule
Semi-superv. & $\mathbf{29.12}$ & $0.31$ & $\mathbf{0.63}$ & $\mathbf{22.96}$ & $\mathbf{0.44}$ & $\mathbf{0.71}$ & $21.49$ & $0.47$ & $0.73$ & $20.70$ & $0.48$ & $\mathbf{0.74}$ \\ 
Semi-superv. \& Adversarial & $29.52$ & $\mathbf{0.32}$ & $\mathbf{0.63}$ & $23.32$ & $0.41$ & $0.70$ & $\mathbf{20.67}$ & $\mathbf{0.48}$ & $\mathbf{0.74}$ & $\mathbf{20.23}$ & $\mathbf{0.49}$ & $\mathbf{0.74}$ \\
\bottomrule
\end{tabular}
\end{center}
\caption{\textbf{Comparison with additional adversarial loss on NYU-CS.}
  Comparison of semi-supervised training with (\emph{Semi-superv. \& Adversarial}) 
  and without an additional adversarial loss (\emph{Semi-superv.}) 
  for different metrics on the NYU-CS dataset.
  Note that these results differ slightly from the results in Tab.~\ref{tab:exp:sota} 
  since the results shown here are from earlier experiments, 
  \ie, based on a previous version of the code base, \etc.
  Best results in boldface.
  }
\label{tab:exp:sup:ganonnyu}
\end{table*}

\begin{table}[t]
\begin{center}
\begin{tabular}{l c c c}
\toprule 
$n$ & \multicolumn{3}{c}{189} \\ 
\cmidrule(rl){2-4}
Metric (see~\S\ref{sec:exp:metrics}) & ME & FS80 & JS80 \\ 
\midrule
Semi-superv. & $\mathbf{27.27}$ & $0.30$ & $\mathbf{0.66}$ \\ 
Semi-superv. \& Adversarial & $27.79$ & $\mathbf{0.32}$ & $0.65$ \\ 
\bottomrule
\end{tabular}
\end{center}
\caption{\textbf{Comparison with additional adversarial loss on MV-hands.}
  Comparison of semi-supervised training with (\emph{Semi-superv. \& Adversarial}) 
  and without (\emph{Semi-superv.}) 
  an additional adversarial loss for different metrics on the MV-hands dataset.
  Best results in boldface.
  }
\label{tab:exp:sup:ganonmvhands}
\end{table}

The decoder $g$ needs to output an image closely resembling 
the image of the second view
since the reconstruction loss is still part of its objective.
Nevertheless, $g$ is enforced to focus more on the overall structure of the image 
through the loss term provided by the discriminator.

Additionally, the discriminator can be improved, and thus provide better feedback,
by conditioning it on additional input, 
as has been described, \eg, in~\cite{Mirza2014arxiv_condgan}.
We can condition it on the input from the first view and/or, 
in case of semi-supervised training, the pose.
In the semi-supervised case, the provided pose information is the estimated pose 
for generated samples and the annotated pose for real samples.

In Tab.~\ref{tab:exp:sup:ganonnyu} and~\ref{tab:exp:sup:ganonmvhands} 
we compare the results when using the additional adversarial term
to the results of semi-supervised training without the adversarial term.
We see that adversarial training can improve the results slightly 
for larger numbers of labeled samples $n$, 
but not in cases where only a small number of samples is labeled.
Moreover, note that we obtained the presented results for the adversarial training 
by tuning hyperparameters separately for different $n$ and taking the best results. 
We found that, for different $n$, different conditioning types and settings for 
$\lambda_{\mathrm{a}}$ worked best.
While for a small number of labeled samples, $n = 100$, 
conditioning the discriminator solely on the input and 
a small weight for the adversarial term ($\lambda_{\mathrm{a}} = 0.01$) yielded best results, 
for larger $n$, conditioning on the pose and a larger weight 
$\lambda_{\mathrm{a}} = 0.1$ had a positive impact on the results.

The results point out that training our method with an additional adversarial loss term 
bears potential to improve results.
However, it appears that a sufficient amount of labeled samples is necessary
so that the discriminator can exploit the pose conditioning and 
provide improved feedback for training the decoder.
Additionally, it requires significant tuning to achieve improved performance.
The semi-supervised approach, as described in~\S\ref{sec:met:semisupervised},
on the other hand, does not require such extensive hyperparameter tuning 
for different amounts of labeled data but 
yields consistently improved performance due to the pose specific latent representation.

\subsection{Additional pre-training baseline}
\label{sec:sup:contextencodercomparison}
One of the reviewers suggested that a more recent pre-training baseline would 
make the paper stronger.
In particular, a comparison to Context Encoders~\cite{Pathak2016cvpr_contextencoders} 
was suggested.
Context Encoders are trained to do inpainting. 
That is, large random contiguous parts of the input image are removed for training and 
the model should learn to inpaint the missing regions based on the context.
The idea is that the model needs to learn to recognize the objects 
in the context in order to accomplish this task.

In Tab.~\ref{tab:sup:unsupervised} we compare the latent representations --
pre-trained by different methods -- based on their predictability for the pose 
(\cf~\S\ref{sec:exp:unsupervised}).
The results show that the representations pre-trained using 
Context Encoders~\cite{Pathak2016cvpr_contextencoders} 
are even less predictive for the pose than the representations learned by autoencoders.

One issue of the Context Encoder baseline is the ``domain gap`` between training and testing 
as has been discussed in~\cite{Zhang2017cvpr_splitbrainautoenc}.
That is, at training time parts of the input are missing, 
while the model is applied to full images at test time.
Moreover, we believe that the main idea of Context Encoders does not really apply to 
pose estimation, where one part of the object does not necessarily contain pose information 
for other parts.

\begin{table}[t]
\begin{center}
\begin{tabular}{l c c c c}
\toprule
Number of samples & 100 & 1,000 & 10,000 & 43,640 \\
\midrule
Context Encoders~\cite{Pathak2016cvpr_contextencoders} & $53.4$ & $53.4$ & $53.3$ & $53.8$ \\
Autoencoder & $48.0$ & $47.2$ & $47.3$ & $47.1$ \\ 
\textbf{PreView (Ours)} & $\mathbf{33.4}$ & $\mathbf{29.6}$ & $\mathbf{29.0}$ & $\mathbf{29.0}$ \\
\bottomrule
\end{tabular}
\end{center}
\caption{
  \textbf{Comparison of different pre-training methods on the NYU-CS dataset.}
  Mean joint error for learning a 
  linear layer on top of the frozen latent representation
  with different numbers of labeled samples $n$.
  Best results in boldface.
  }
\label{tab:sup:unsupervised}
\end{table}

\begin{figure*}[t]
  \centering
  \begin{subfigure}{0.48\textwidth}
    \includegraphics[width=\textwidth]{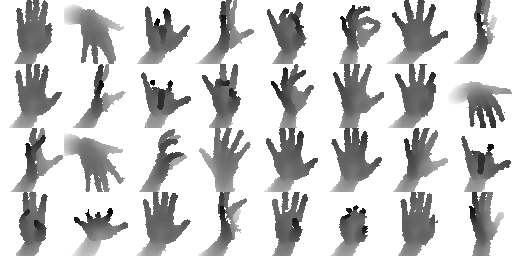}
    \caption{Input view}
  \end{subfigure}
  \quad
  \begin{subfigure}{0.48\textwidth}
    \includegraphics[width=\textwidth]{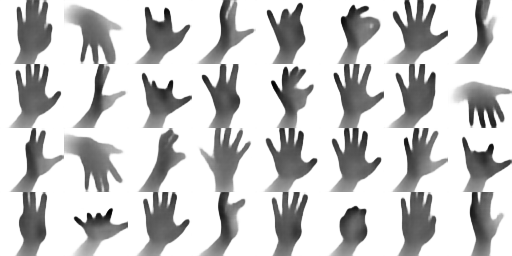}
    \caption{Input reconstruction}
  \end{subfigure}
  \\
  \begin{subfigure}{0.48\textwidth}
    \includegraphics[width=\textwidth]{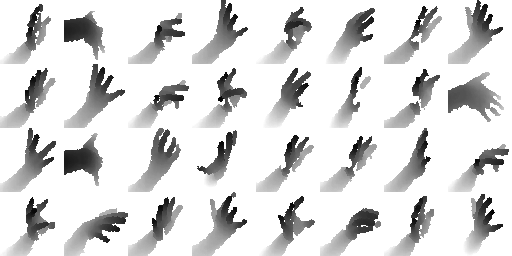}
    \caption{Different view}
  \end{subfigure}
  \quad
  \begin{subfigure}{0.48\textwidth}
    \includegraphics[width=\textwidth]{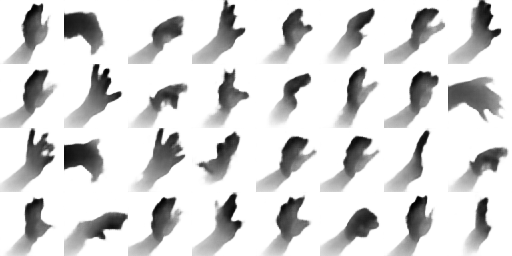}
    \caption{Prediction for different view}
  \end{subfigure}
  \caption{\textbf{Input reconstruction vs. different view prediction.}
  Examples for generated views from the NYU validation set. 
  Input view~(a), reconstructions generated by the autoencoder~(b), 
  images from a different view~(c), 
  and the corresponding predictions from our method~(d).
  Images from~(a)-(d) with same grid index are corresponding.
  Visually, the autoencoder's input reconstructions resemble the input more closely
  than the predictions of our method match the different view.
  However, the latent representation learned by our method is much more predictive for the pose 
  (\cf, results in~\S\ref{sec:exp:unsupervised}).}
  \label{fig:predictedviewsnyu}
\end{figure*}

\subsection{Results for Model Selection on MV-hands}
\label{sec:sup:novalmvhands}
In~\S\ref{sec:exp:semisupervised} we compare the results on the MV-hands dataset 
when training on all data available 
for training, \ie, using the 100 samples from the validation set for training, 
and omitting any model selection 
like early stopping or other hyperparameter optimizations.
For comparison, in Tab.~\ref{tab:exp:sup:semisupervised:ablation:icg:189} 
we provide the results when employing the 100 validation samples for 
model selection during training, \ie, early stopping.

\begin{table}[t]
\begin{center}
\begin{tabular}{l c c c}
\toprule 
$n$ & \multicolumn{3}{c}{189} \\ 
\cmidrule(rl){2-4}
Metric (see~\S\ref{sec:exp:metrics}) & ME & FS80 & JS80 \\ 
\midrule
DeepPrior++~\cite{Oberweger2017iccvw_deeppriorpp} & $36.56$ & $0.17$ & $0.54$ \\ 
\midrule
Supervised & $30.13$ & $0.25$ & $0.62$ \\ 
Semi-superv. Autoenc. & $28.51$ & $\mathbf{0.30}$ & $\mathbf{0.64}$ \\ 
\textbf{Semi-superv. PreView (Ours)} & $\mathbf{28.38}$ & $\mathbf{0.30}$ & $\mathbf{0.64}$ \\ 
\bottomrule
\end{tabular}
\end{center}
\caption{\textbf{Comparison to the state-of-the-art and ablation experiments.}
  Results for different metrics on the MV-hands dataset.
  Best results in boldface.
  }
\label{tab:exp:sup:semisupervised:ablation:icg:189}
\end{table}

\subsection{Predicted views}
\label{sup:sec:predictedviews}

\begin{figure*}[t]
  \centering
  \begin{subfigure}{0.48\textwidth}
    \includegraphics[width=\textwidth]{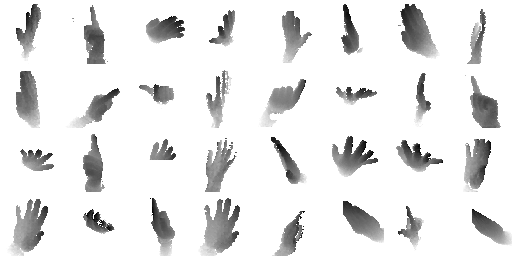}
    \caption{Different view}
  \end{subfigure}
  \quad
  \begin{subfigure}{0.48\textwidth}
    \includegraphics[width=\textwidth]{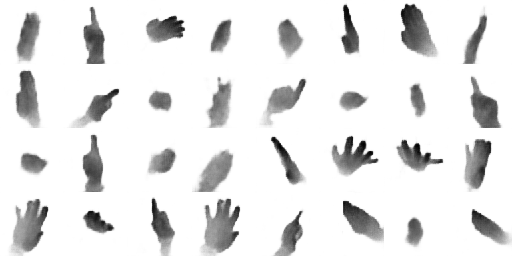}
    \caption{Prediction for different view}
  \end{subfigure}
  \caption{\textbf{View prediction examples on MV-hands data.}
  Target view~(a), \ie, ground truth images of the different view 
  and the corresponding predictions from our method~(b).
  Images with same grid index are corresponding.}
  \label{fig:predictedviewsmvhands}
\end{figure*}

In Fig.~\ref{fig:predictedviewsnyu} 
we compare output images of 
input reconstruction (autoencoder) and view prediction (PreView), respectively.
We can observe that the reconstructions of the input are cleaner 
(\eg for the fingers) than the predictions for different views.
Obviously, reconstructing the input is an easier task than predicting a different view.
More importantly, input reconstruction can 
be performed without knowledge about the pose, as the results 
in~\S\ref{sec:exp:unsupervised} suggest. 
Predicting different views, on the other hand, is a harder task but reveals pose information.
In Fig.~\ref{fig:predictedviewsmvhands} we show view prediction examples 
on the MV-hands dataset.
Altogether, these results underline that our latent representation is predictive 
for the different view as well as the pose.

\vspace{0.7em}

\subsection{Neuron activations}
\label{sup:sec:neuronactivations}
\begin{figure*}[t]
  \centering
  \begin{subfigure}{0.48\linewidth}
    \includegraphics[width=\textwidth]{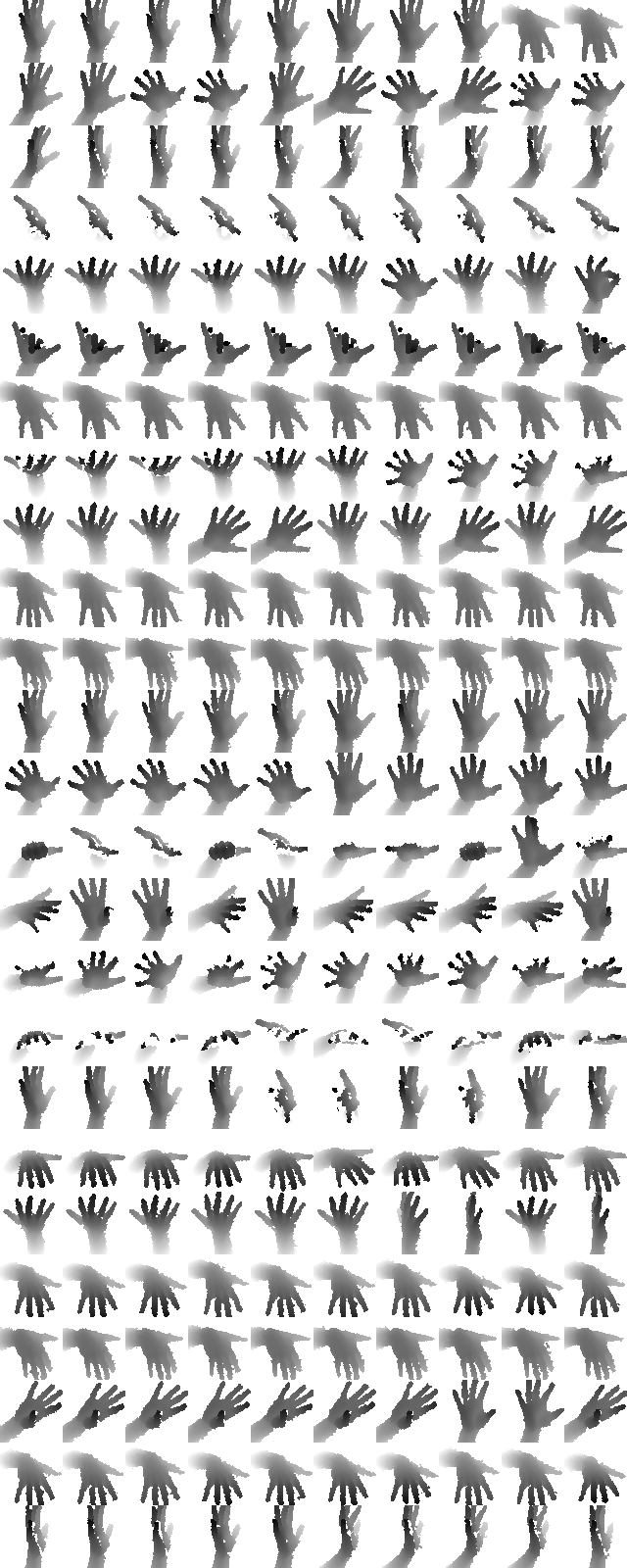}
  \end{subfigure}
  \quad
  \begin{subfigure}{0.48\linewidth}
    \includegraphics[width=\textwidth]{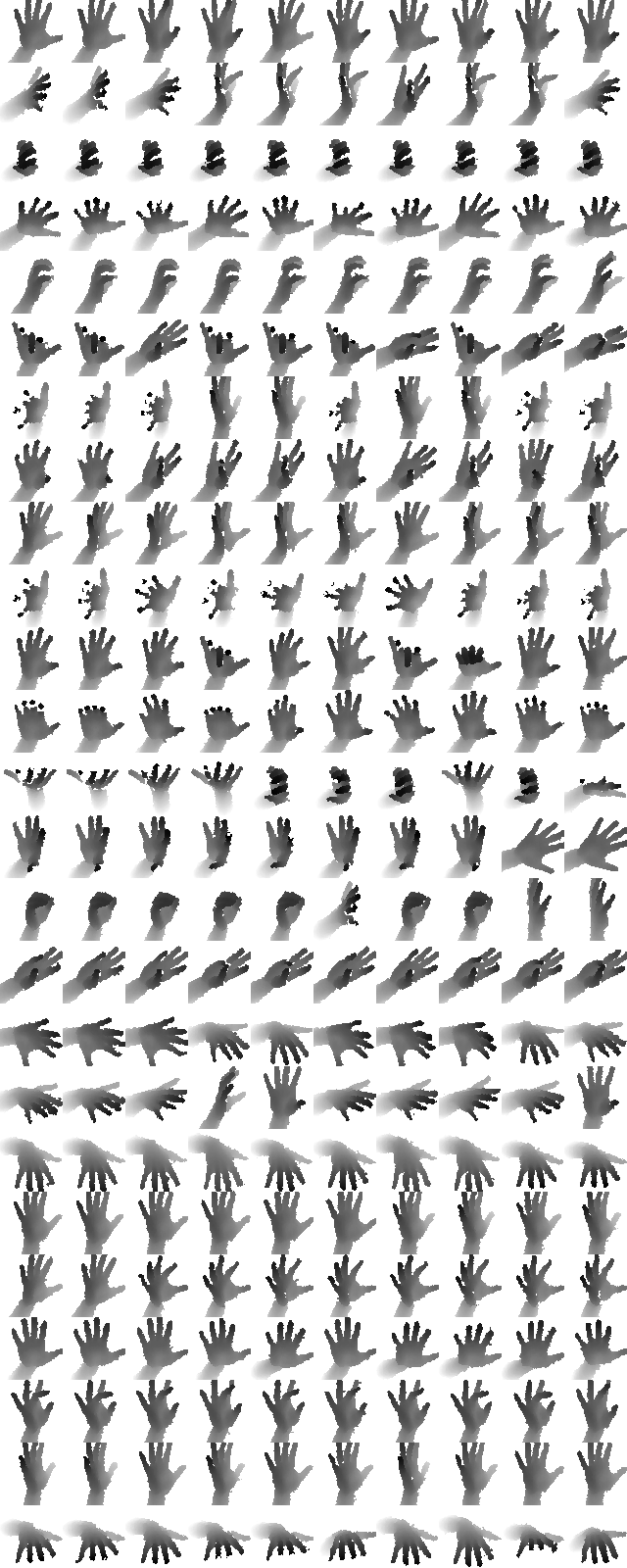}
  \end{subfigure}
  \caption{\textbf{Most activating samples.}
  Each row on the left and right side shows the ten samples from the validation set, 
  which activate the same neuron in the learned latent representation most.
  Note, that we randomly perturbed detections to verify the robustness of our method.
  Hence, sometimes parts of the hand are cut off in the crops.}
  \label{fig:topactivated}
\end{figure*}

To investigate what each neuron in the latent space has learned,
we search for the samples from the validation set, which activate a single neuron most.
Fig.~\ref{fig:topactivated} shows these samples for each neuron.
We find that many of the neurons are activated most for very specific poses.
That is, the samples, which activate a neuron most, show similar poses.

\vspace{0.7em}

\subsection{Nearest neighbors}
\label{sup:sec:nearestneighbors}
\begin{figure}[t]
  \centering
  \includegraphics[width=\linewidth]{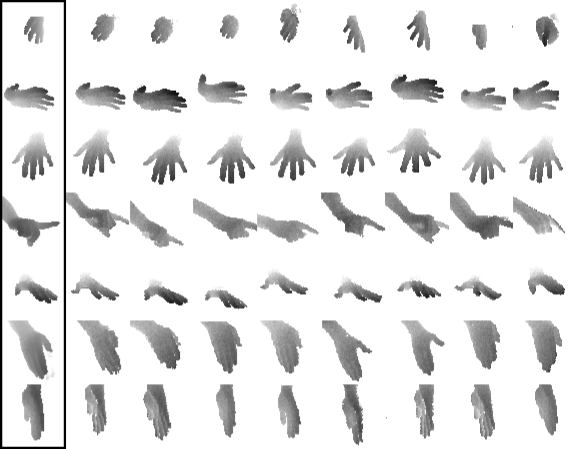}
  \caption[]{\textbf{Nearest neighbors in latent space.}
  Nearest neighbors from training set of the MV-hands dataset for 
  query samples from validation set.
  Query images are shown in the marked, leftmost column, 
  the remaining eight columns are the respective nearest neighbors.}
  \label{fig:exp:sup:nearestneighbors}
\end{figure}

Similar to what we show in~\S\ref{sec:exp:unsupervised} for the NYU dataset,
in Fig.~\ref{fig:exp:sup:nearestneighbors}, we show nearest neighbors for the MV-hands dataset.
That is, given a query image from the validation set, we find the closest samples
from the training set according to the Euclidean distance in the learned latent 
representation space.
Again, the nearest neighbors in the latent space exhibit very similar poses.

\end{document}